\documentclass[letterpaper]{article} 
\usepackage{aaai2026}  
\usepackage{times}  
\usepackage{helvet}  
\usepackage{courier}  
\usepackage[hyphens]{url}  
\usepackage{graphicx} 
\usepackage{natbib}  
\usepackage{caption} 
\usepackage{algorithm}
\usepackage{algorithmic}
\usepackage{algorithm}
\usepackage{algorithmic}
\usepackage{amsmath}
\usepackage{enumitem}
\usepackage{tabularray}
\usepackage{tikz}
\usepackage{xcolor}
\usepackage{colortbl}
\usepackage{amsmath}
\usepackage{graphicx}
\usepackage{array}
\usetikzlibrary{shapes, arrows.meta}
\definecolor{darkgreen}{RGB}{0,100,0}
\usepackage[table]{xcolor}
\usepackage{colortbl}
\usepackage{booktabs}
\usepackage{enumitem}

\usepackage{amsfonts}
\usepackage{newfloat}
\usepackage{listings}
\DeclareCaptionStyle{ruled}{labelfont=normalfont,labelsep=colon,strut=off} 
\floatstyle{ruled}
\newfloat{listing}{tb}{lst}{}
\floatname{listing}{Listing}
\title{CausalTrace: A Neurosymbolic Causal Analysis Agent for Smart Manufacturing}
\author{
    Chathurangi Shyalika\textsuperscript{\rm 1}, Aryaman Sharma\textsuperscript{\rm 1}, Fadi El Kalach\textsuperscript{\rm 2}, Utkarshani Jaimini\textsuperscript{\rm 3}\footnote{Work done while at {University of South Carolina}}, \\Cory Henson\textsuperscript{\rm 4}, Ramy Harik\textsuperscript{\rm 2}, Amit Sheth\textsuperscript{\rm 1}}

\affiliations{
    \textsuperscript{\rm 1}Artificial Intelligence Institute, University of South Carolina\\
    \textsuperscript{\rm 2}Department of Automotive Engineering, Clemson University \\
    \textsuperscript{\rm 3}Stride Lab, University of Michigan, Dearborn
\\
    \textsuperscript{\rm 4}Bosch Center for Artificial Intelligence, Pittsburgh
  
}

\usepackage{bibentry}

\begin{document}

\maketitle
\begin{abstract}
Modern manufacturing environments demand not only accurate predictions but also interpretable insights to process anomalies, root causes, and potential interventions. Existing AI systems often function as isolated black boxes, lacking the seamless integration of prediction, explanation, and causal reasoning required for a unified decision-support solution. This fragmentation limits their trustworthiness and practical utility in high-stakes industrial environments. In this work, we present CausalTrace, a neurosymbolic causal analysis module integrated into the SmartPilot industrial CoPilot. CausalTrace performs data-driven causal analysis enriched by industrial ontologies and knowledge graphs, including advanced functions such as causal discovery, counterfactual reasoning, and root cause analysis (RCA). It supports real-time operator interaction and is designed to complement existing agents by offering transparent, explainable decision support. We conducted a comprehensive evaluation of CausalTrace using multiple causal assessment methods and the C3AN framework (i.e. Custom, Compact, Composite AI with Neurosymbolic Integration), which spans principles of robustness, intelligence, and trustworthiness. In an academic rocket assembly testbed, CausalTrace achieved substantial agreement with domain experts (ROUGE-1: 0.91 in ontology QA) and strong RCA performance (MAP@3: 94\%, PR@2: 97\%, MRR: 0.92, Jaccard: 0.92). It also attained 4.59/5 in the C3AN evaluation, demonstrating precision and reliability for live deployment.



\end{abstract}


\section{Introduction}
Manufacturing is entering an era of hyperautonomous operations, powered by advances in AI-enabled sensing, control, and decision support~\cite{hasan2026hyperautomation}.
As production systems grow in complexity, there is an urgent need for solutions that integrate causal reasoning with human-interpretable decision support to ensure adaptive and trustworthy operation in safety-critical environments.

While machine learning models have demonstrated impressive results in demand forecasting and anomaly detection, their lack of interpretability remains a significant barrier to adoption in high-stakes industrial settings. These black-box models often fail to meet the practical needs of shop-floor operators and subject matter experts (SMEs), who require not only accurate predictions but also actionable and understandable insights into system behavior~\cite{singh2024fundamental}. This has fueled the demand for AI systems that can explain why events occur, identify underlying causes, and offer counterfactual reasoning to support 'what if' analysis. These capabilities are essential for real-time decision-making, proactive maintenance, and effective human-machine collaboration in production workflows~\cite{acharya2025agentic, boskabadi2025industrial}. Yet, existing AI solutions rarely integrate these functionalities into a unified, deployable framework, limiting their utility in live industrial environments.

To address this gap, we present \textbf{CausalTrace}, a neurosymbolic causal analysis agent embedded in our prior work, \textit{SmartPilot}\footnote{Code: https://github.com/ChathurangiShyalika/SmartPilot, \\Demo: https://smartpilot.my.canva.site} industrial CoPilot~\cite{shyalika2025smartpilot_1, shyalika2025smartpilot_2}. SmartPilot builds on the vision of \textbf{C3AN}—\textit{Custom, Compact, Composite AI with Neurosymbolic Integration} \cite{c3an_new}—\,an AI paradigm emphasizing robustness, intelligence, and trust of AI systems. \textit{Custom} in C3AN refers to domain-specific data and workflows and \textit{Compact} refers to efficient models deployable across infrastructures, including edge devices. \textit{Composite} refers to the modular orchestration of specialized AI components- neural, symbolic, and decision modules- into a unified system. \textit{Neurosymbolic} denotes the integration of neural learning and symbolic reasoning within each module, enabling transparent, knowledge-aligned, and explainable decision-making. Key contributions of this work are:
\vspace{-1mm}
\begin{itemize}
    \item Design and implement CausalTrace, a neurosymbolic causal analysis agent capable of real-time causal discovery, root cause analysis (RCA), causal effect estimation, and counterfactual reasoning.
    \item Integrate data-driven causal methods with structured knowledge (ontologies, knowledge graphs), semantic user interfaces, and evaluation pipelines into neurosymbolic agent workflows, enabling human-interpretable reasoning and a deployable decision-support system.
    \item Develop a comprehensive evaluation methodology grounded in the C3AN framework, incorporating a suite of techniques to ensure robustness, intelligence, and trustworthiness. Conduct causal assessments including counterfactual effect estimation, RCA validation, and comparisons with baseline and ablation variants.
    \item Demonstrate the deployment and practical utility of CausalTrace on an academic rocket assembly testbed, highlighting its readiness for real-world manufacturing environments.
\end{itemize}

\vspace{-4mm}
\section{Related Work}

\subsubsection{Causal Reasoning in Industrial AI}
Research on causal reasoning in industrial settings spans both symbolic and data-driven approaches. Early systems employ causal models rooted in qualitative physics and control theory to support fault localization and diagnostic reasoning \cite{bandekar1989causal}. Other efforts introduce planning-based frameworks that embed causal reasoning into multi-robot coordination and factory automation, enabling optimal plan reuse and failure diagnosis in dynamic production environments \cite{erdem2012causality}. More recently, causal discovery techniques such as nonparametric entropy methods and information-geometric inference have been used to reconstruct causal networks directly from process data, guiding performance prediction and process optimization in complex industrial workflows \cite{sun2024reconstructing}.

\vspace{-1mm}
\subsubsection{Neurosymbolic and Knowledge-Infused Systems}
 In safety-critical domains like manufacturing and cybersecurity, neurosymbolic systems have shown promise in improving explainability and behavior under uncertainty by combining neural networks with explicit knowledge graphs \cite{piplai2023knowledge}. In scientific domains such as geochemistry, symbolic rules extracted from expert literature via LLMs have been integrated with learning models to guide predictions and improve domain alignment \cite{chen2025integrating}. 

\vspace{-1mm}
\subsubsection{Agentic AI and Multiagent Architectures in Industry}
Agentic AI represents a paradigm shift toward autonomous systems capable of goal-driven, adaptive decision-making. Surveys of Agentic AI emphasize its core attributes: autonomy, proactivity, learning, and reactivity as enablers for complex real-world applications \cite{acharya2025agentic}. Agentic systems have been integrated with digital twins to support real-time monitoring and control of nonlinear systems \cite{kusiak2025agentic, boskabadi2025industrial}. Recent frameworks such as Intelligent Design 4.0 \cite{jiang2025intelligent}, intent-based industrial automation \cite{romero2025agentic}, and multi-agent coordination architectures \cite{panigrahy2025multi} propose layered systems where LLM-powered agents decompose high-level goals and orchestrate downstream execution across various functions. 

\begin{figure}[!htb]
  \centering
\includegraphics[width=0.999\linewidth]{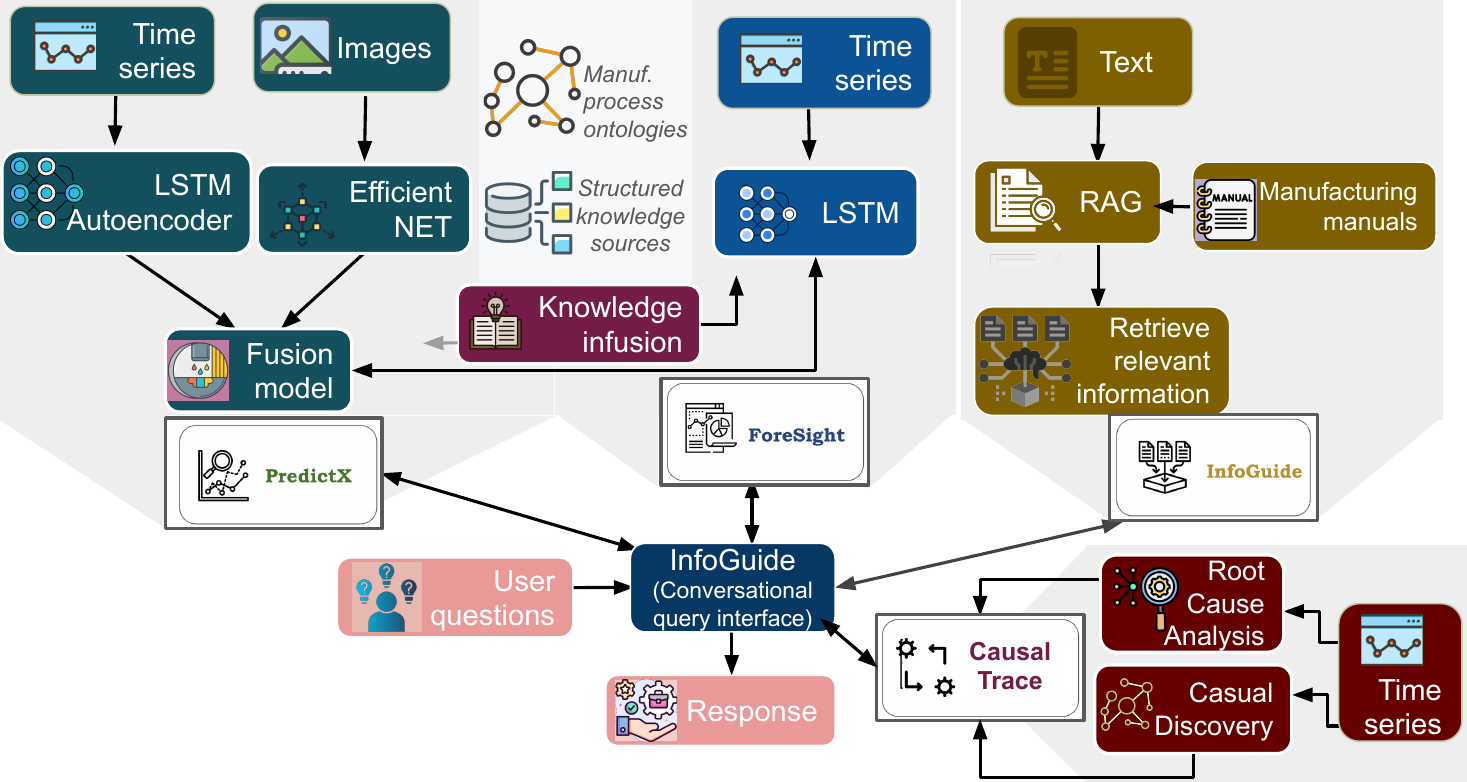}
\caption{Multi-agent architecture of SmartPilot}
  \label{fig:overall_architecture}
  \vspace{-6mm}
\end{figure}

\vspace{-1mm}
\subsubsection{Summary and Motivation}
While prior work in causal reasoning, neurosymbolic AI, and agentic systems has made meaningful strides, current solutions often fall short in integration, interactivity, and operational readiness. Symbolic methods struggle to scale; neural models lack transparency; and agentic systems lack semantic grounding and are not designed for operator-in-the-loop decision support. These gaps hinder adoption in complex industrial environments where explainability, adaptability, and human collaboration are critical. SmartPilot is built to address these limitations holistically. By fusing causality, neurosymbolic methods, and agentic AI, it lays a practical and extensible foundation for intelligent manufacturing.

\vspace{-3mm}
\section{System Design}
\vspace{-0.5mm}
\subsection{SmartPilot Architecture} 
\vspace{-1mm}The earlier version of the SmartPilot system adopts a multi-agent architecture, comprising three specialized agents that collectively support predictive analytics and operator assistance. The algorithmic details of these agents are available at \cite{shyalika2025nsf, shyalika2025smartpilot_1, shyalika2025smartpilot_2}. PredictX agent specializes in anomaly prediction using multimodal sensor data (time series and images). It leverages decision-level fusion enhanced by transfer learning and knowledge-infused learning (via process ontologies) in predicting anomalies. ForeSight agent employs a Knowledge-Infused LSTM (KIL-LSTM) model to predict throughput across production cycles. It integrates time series data with domain priors to improve long-term forecasting. InfoGuide is a question-answering agent that provides real-time responses to operational, safety, maintenance, and troubleshooting queries. It uses retrieval-augmented generation over curated manufacturing manuals and integrates with other agents to address questions related to anomaly detection, production forecasting, and causal reasoning. SmartPilot’s natural language understanding and explanation are powered by \texttt{LLaMA3-70B-8192} model.

\vspace{-2mm}
\subsection{Integrating the CausalTrace Agent}
\vspace{-0.5mm}

\begin{figure*}[!htb]
  \centering
\includegraphics[width=0.999\linewidth]{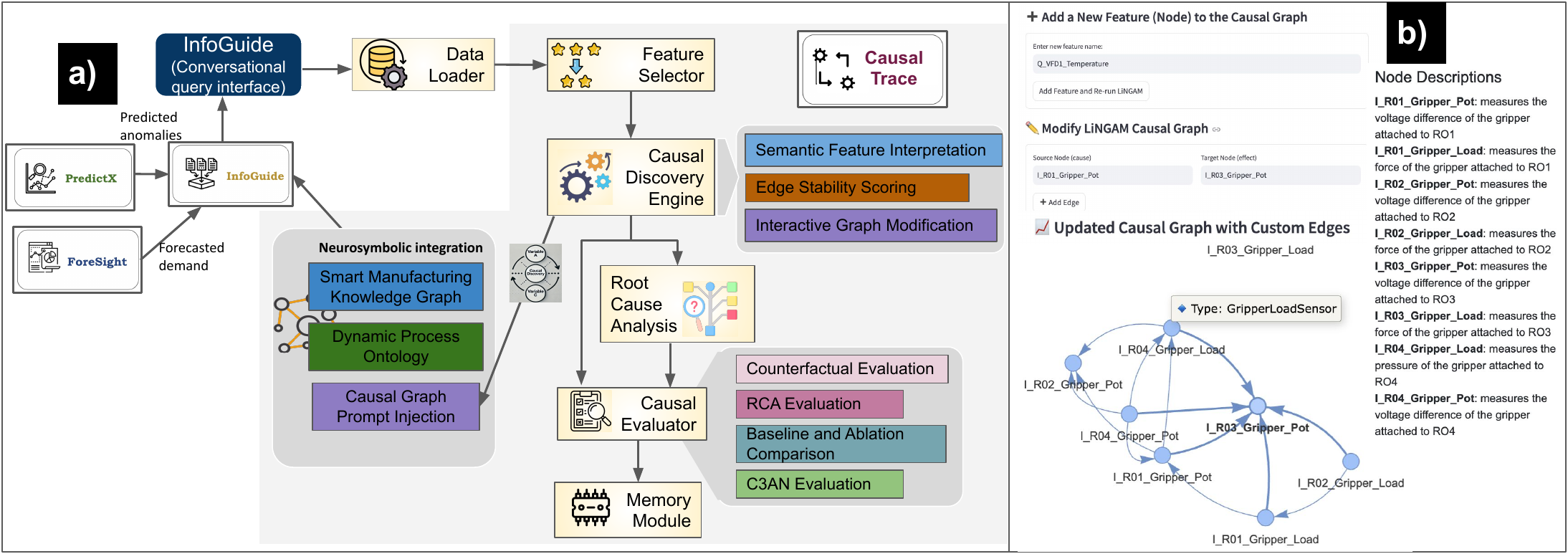}
\vspace{-6mm}
\caption{
(a) \textit{Architecture of CausalTrace agent}:  PredictX provides predicted anomalies, ForeSight offers forecasted demand, both feeding into InfoGuide. Data flows through Data Loader and Feature Selector before entering the Causal Discovery Engine. Discovered causal graphs support Root Cause Analysis. Causal graphs and RCA results are validated in the Causal Evaluator. InfoGuide responses are enhanced via neurosymbolic integration, combining a manufacturing knowledge graph, process ontology, and causal graph prompts. Final results are stored in the Memory Module. (b) \textit{Interactive User Interface}: Causal graph visualization and allows causal graph editing.
Enhances graphs with ontology-derived descriptions, types, and units. Tooltip metadata is enriched using a smart manufacturing knowledge graph.
}
  \label{fig:overall_causal_architecture}
  \vspace{-6mm}
\end{figure*}

The CausalTrace agent extends SmartPilot’s capabilities beyond prediction to include explanation, attribution, and simulation needed for efficient RCA. This agent is tightly coupled into SmartPilot (Figure \ref{fig:overall_architecture}) and supports real-time and historical data analysis through an interactive interface. The causal analysis pipeline (Figure \ref{fig:overall_causal_architecture}) includes the following components and features.

\vspace{-1mm}

\subsubsection{Data Loader and Feature Selector.}
SmartPilot’s Data Loader supports two modes of data ingestion: direct connection to Programmable Logic Controllers (PLCs) for real-time streaming, or batch upload of historical sensor data for offline analysis. Once data is ingested, the Feature Selector enables the selection of data-driven feature selection methods or manual selection based on domain expertise, allowing operators to prioritize variables most relevant to the analysis.

\vspace{-1mm}
\subsubsection{Causal Discovery Engine.} 
The engine supports Independent Component Analysis–based Linear Non-Gaussian Acyclic Model (ICA-based LiNGAM)  \cite{shimizu2006linear} and Differentiable Causal Discovery with Attention (DiffAN) \cite{sanchez2022diffusion} to construct Directed Acyclic Graphs (DAGs) from multivariate sensor data. To enhance reliability, we integrate bootstrap-based edge stability analysis (Algorithm 1) directly into the discovery workflow. Causal graphs are repeatedly estimated on bootstrap-resampled datasets, and edge stability scores computed from their occurrence frequency are used to down-weight or remove low-confidence edges (Algorithm 1). This yields more trustworthy graphs while preserving meaningful causal structure. For each retained edge, the total causal effect is computed (Algorithm 2) to quantify both direct and indirect influences between variables, and results are visualized in the user interface for interactive exploration.

\begin{algorithm}[!ht]
\caption*{\textbf{Algorithm 1: Bootstrap-Based Edge Stability Analysis}}
\begin{flushleft}
\textbf{Input:} Dataset $\mathcal{D}$, number of bootstrap samples $N$, causal discovery method $M$\ \\
\textbf{Output:} Stability score $s$ for each edge $A \rightarrow B$
\end{flushleft}
\vspace{-0.5em}
\begin{algorithmic}[1]
\STATE Initialize list $\mathcal{W}{A \rightarrow B} = []$ for each $A \rightarrow B$
\FOR{$i = 1$ to $N$}
\STATE Resample dataset $\mathcal{D}^{(i)}$ with replacement from $\mathcal{D}$
\STATE Run $M$ on $\mathcal{D}^{(i)}$ to estimate causal graph $\mathcal{G}^{(i)}$
\FOR{each directed edge $A \rightarrow B$ in $\mathcal{G}^{(i)}$}
\STATE Record causal strength $w_i$ of $A \rightarrow B$
\STATE Append $w_i$ to $\mathcal{W}{A \rightarrow B}$
\ENDFOR
\ENDFOR
\FOR{each edge $A \rightarrow B$ with samples $\mathcal{W}{A \rightarrow B} = {w_1, \dots, w_N}$}
\STATE Compute mean strength: $\bar{w} = \frac{1}{N} \sum{i=1}^{N} w_i$
\STATE Compute standard deviation: \\
$\sigma = \sqrt{\frac{1}{N} \sum_{i=1}^{N} (w_i - \bar{w})^2}$
\STATE Compute stability score: $s = \frac{1}{1 + \sigma}$
\ENDFOR
\STATE \textbf{Edge Interpretation:}
\begin{itemize}
\item $s \geq 0.9$: Very strong and stable
\item $0.8 \leq s < 0.9$: Reliable
\item $0.6 \leq s < 0.8$: Moderately stable (use with caution)
\item $s < 0.6$: Unstable (typically excluded)
\end{itemize}
\STATE Retain edges with $s \geq 0.6$ for the final causal graph
\end{algorithmic}
\vspace{-1mm}
\end{algorithm}

\begin{algorithm}[!ht]
\caption*{\textbf{Algorithm 2: Total Causal Effect Computation}}
\begin{flushleft}
\textbf{Input:} Causal effect matrix $\mathbf{B}$, identity matrix $\mathbf{I}$\\
\textbf{Output:} Total causal effect matrix $\mathbf{T} = (\mathbf{I} - \mathbf{B})^{-1}$
\end{flushleft}
\vspace{-0.5em}
\begin{algorithmic}[1]
\STATE Start with a structural equation model (SEM): $\mathbf{X} = \mathbf{B} \mathbf{X} + \boldsymbol{\varepsilon}$, where $\mathbf{X}$ is a vector of observed variables and $\boldsymbol{\varepsilon}$ is noise.
\STATE Rearranged to: $(\mathbf{I} - \mathbf{B}) \mathbf{X} = \boldsymbol{\varepsilon}$
\STATE Solve for $\mathbf{X}$ using: $\mathbf{X} = (\mathbf{I} - \mathbf{B})^{-1} \boldsymbol{\varepsilon}$.
\STATE Therefore, the total effect matrix is: $\mathbf{T} = (\mathbf{I} - \mathbf{B})^{-1}$.
\STATE Each entry $T_{ij}$ captures total effect of variable $j$ on variable $i$
\STATE This can be decomposed as $\mathbf{T} = \mathbf{I} + \mathbf{B} + \mathbf{B}^2 + \mathbf{B}^3 + \cdots$.
\STATE Interpretation:
\begin{itemize}
    \item $\mathbf{I}$: Self influence (identity effect),
    \item $\mathbf{B}$: Direct effects (1-hop causal influence),
    \item $\mathbf{B}^k$: Indirect effects through $k$-hop paths.
\end{itemize}
\STATE Series expansion valid when graph is acyclic or spectral radius of $\mathbf{B} < 1$
\STATE $\Rightarrow$ Total causal effect aggregates all direct and mediated influences
\end{algorithmic}
\vspace{-1mm}
\end{algorithm}

\vspace{-1mm}
\subsubsection{Root Cause Analysis.} 
For each anomaly, the RCA module combines expert-defined, cycle-aware sensor tolerance ranges with the learned causal topology to generate a ranked list of candidate root causes, based on causal path effect strengths and sensor deviation analysis.

\vspace{-1mm}
\subsubsection{Neurosymbolic Integration.} SmartPilot adopts a multi-layered neurosymbolic integration framework that tightly couples structured knowledge with data-driven reasoning. The system leverages a \textit{smart manufacturing knowledge graph} encoded in RDF to represent entities such as sensors, machines, parts, and anomalies. This knowledge is injected into the reasoning layer using \texttt{rdflib}, enriching InfoGuide’s responses with semantic context. A \textit{dynamic process ontology}\footnote{https://github.com/revathyramanan/Dynamic-Process-Ontology}, implemented in Neo4j, encodes process semantics, variable relationships, and operational constraints. Real-time \texttt{Cypher} queries enable on-demand retrieval of explanations, tolerance ranges, and sensor-function mappings. The \textit{causal graph prompt injection} mechanism serializes the total causal effect matrix from the causal discovery step and passes it into LLM prompts. This allows the system to generate explanations and reasoning that are grounded in the structure of the causal graph.

\vspace{-1.5mm}
\subsubsection{Interactive User Interface.} 

The interface visualizes causal graphs with rich semantic metadata derived from the process ontology and the knowledge graph. Each node is annotated with descriptions, types, units, and anomaly relevance, and tool tips display this semantic information for interactive exploration. CausalTrace is integrated with InfoGuide’s conversational query interface, enabling operators to ask natural language questions about graph structure, causal reasoning, and root cause analysis. Example queries are shown in Figure~\ref{fig:q_types}, with a complete set of competency questions provided in Appendix A. Users can also interactively modify the discovered causal graph by adding or removing nodes and edges; all edits are validated against the ontology to ensure semantic consistency, allowing domain experts to refine and adapt the graph in real time.

\vspace{-1mm}
\subsubsection{Memory Module.} This enables persistent, context-aware reasoning by storing and retrieving information across sessions. It consists of three components: \textit{episodic memory}, which logs time-stamped interactions such as causal discovery and RCA runs for longitudinal tracking; \textit{semantic memory}, which stores structured annotations of sensors and entities to support enriched, context-aware explanations; and \textit{procedural memory}, which retains user preferences (e.g., chosen algorithms or display settings) to enable personalized interactions. These memory traces are stored in JSON format and are injected into InfoGuide’s natural language responses to ensure continuity and context-aware dialogue.

\begin{figure}[!ht]
\centering
\begin{tikzpicture}[
    var/.style={circle, draw=black, minimum size=0.8cm, font=\scriptsize, inner sep=1pt},
    querybox/.style={rectangle, draw, rounded corners, fill=gray!10, align=left, text width=5.4cm, font=\scriptsize, inner sep=4pt},
    thickarrow/.style={->, thick}
  ]

  \node[var, fill=blue!20] (A) at (0,0) {A};
  \node[var, fill=green!20] (B) at (0,2) {B};
  \node[var, fill=red!20] (C) at (0,4) {C};
\node[var, fill=orange!20] (D) at (-1.3,1) {D};

  \draw[thickarrow] (A) -- (B);
  \draw[thickarrow] (B) -- (C);
  \draw[thickarrow] (D) -- (B);

  \node[querybox, fill=blue!10] at (3.5,4) {
    \textbf{Causal Graph Structure} \\
    -- Is there a causal relation between A and B? \\
    -- What are the causal parents of C?
  };

  \node[querybox, fill=green!10] at (3.5,2.4) {
    \textbf{Causal Reasoning} \\
    -- If the value of A is set to value $x$, what would be its effect on B? \\
    -- What is the strength of the causal relation between B and C?
  };

  \node[querybox, fill=orange!10] at (3.5,0.6) {
    \textbf{Root Cause Analysis} \\
    -- What is the strongest cause of the anomalous value of variable B? \\
    -- Why is D not likely a root cause of B?
  };

\end{tikzpicture}
\vspace{-2mm}
\caption{Illustration of a structured causal graph with categorized query types: structure, reasoning, and RCA.}
\label{fig:q_types}
\vspace{-6mm}
\end{figure}
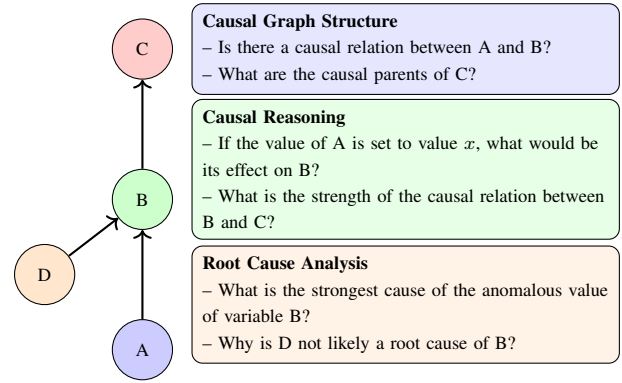

\vspace{-2mm}
\section{Evaluation and Results}

\subsubsection{Dataset Description.}
We use the publicly available manufacturing dataset generated by the Future Factories (FF) Lab at the McNair Aerospace Research Center, University of South Carolina~\cite{harik2024analog}. This dataset captures synchronized sensor and image data from a prototype rocket assembly pipeline designed to emulate industrial-grade manufacturing processes. It contains 166K records sampled at 1.95 Hz over 30 hours, covering 285 complete rocket assembly–disassembly cycles. Each cycle is segmented into 21 distinct operational states. The dataset includes time-series measurements from potentiometers, load cells, drive temperatures, and robot kinematics. The annotated version used in this work \cite{shyalika2025nsf}\footnote{https://github.com/ChathurangiShyalika/NSF-MAP} includes cycle state labels along with ground-truth anomaly types. These anomalies indicate missing rocket components and include six types, such as \texttt{NoNose}, \texttt{NoBody2}, and combined cases like \texttt{NoBody2,NoBody1}. In the evaluation, causal discovery was performed using ICA-based LiNGAM and DiffAN on eight key variables identified through XGBoost feature selection and validated by domain experts. LiNGAM produced 20 directed causal edges, while DiffAN yielded 15. The subsections below detail the evaluation methods conducted on these causal graphs.

\vspace{-1mm}
\subsubsection{Counterfactual Effect Computation.} 


CausalTrace provides an interactive module for validating causal links via counterfactual effect computation. From the user interface, users can select an edge $A \rightarrow B$, apply an intervention on $A$, and compare the predicted change in $B$ (from the learned total effect; Algorithm~3) with observed changes in held-out data. This enables direct plausibility checks and graph refinement based on domain knowledge.

\vspace{-1mm}
\begin{algorithm}[!ht]
\caption*{\textbf{Algorithm 3:Counterfactual Validation of Causal Effects}}
\begin{flushleft}
\textbf{Input:} Dataset $\mathcal{D}$, total effect matrix $\tau$, tolerance $\epsilon$, effect threshold $\delta$ \\
\textbf{Output:} Validation table with $(A, B, \tau_{A \rightarrow B}, \Delta B_{\text{pred}}, \Delta B_{\text{obs}}, \text{Error})$ for each valid causal pair $A \rightarrow B$
\end{flushleft}
\vspace{-0.5em}
\begin{algorithmic}[1]

\FOR{each variable pair $(A, B)$ where $A \ne B$ and $|\tau_{A \rightarrow B}| > \delta$}
    \STATE Estimate baseline ($a_1$) and intervention ($a_2$) values. If unspecified, compute:
\[
a_1 = Q_1(A) \text{ (1st quartile)}, \quad a_2 = Q_3(A) \text{ (3rd quartile)}
\]
    \vspace{-4mm}
    \STATE Compute predicted change using total effect:
    \vspace{-2mm}
    \[
    \Delta B_{\text{pred}} = (a_2 - a_1) \cdot \tau_{A \rightarrow B}
    \]
    \vspace{-4mm}
    \STATE Extract subsets of $\mathcal{D}$:\\
      - Baseline group: rows where $A \approx a_1$ within  $\epsilon$ \\
      - Counterfactual group: rows where $A \approx a_2$ within $\epsilon$

    \STATE Compute observed change in $B$:
    \vspace{-2mm}
    \[
    \Delta B_{\text{obs}} = \mathbb{E}[B \mid A \approx a_2] - \mathbb{E}[B \mid A \approx a_1]
    \]
    \vspace{-4mm}
    \STATE Compute absolute error: $\text{Error} = \left| \Delta B_{\text{pred}} - \Delta B_{\text{obs}} \right|$

    \STATE Record result: $(A, B, \tau_{A \rightarrow B}, \Delta B_{\text{pred}}, \Delta B_{\text{obs}}, \text{Error})$
\ENDFOR

\STATE Compile results into a table for global causal graph assessment

\STATE \textbf{Interpretation:}
\begin{itemize}
    \item Small error $\Rightarrow$ Causal effect from $A$ to $B$ is well-supported by data
    \item Large error $\Rightarrow$ Possible misspecification or unobserved confounding
\end{itemize}

\end{algorithmic}
\vspace{-1mm}
\end{algorithm}

\subsubsection{Root Cause Analysis Evaluation.} We evaluated the RCA module on anomaly events from the FF dataset with known fault cases. For each anomaly, CausalTrace produced a ranked list of candidate root causes using causal path significance and sensor deviation analysis. Here, $K$ denotes the cut-off rank that is, the number of top predictions considered when computing the metrics. Results showed close agreement with expert-defined causes, achieving Mean Average Precision@K (MAP@3) of 94\%, Precision@K (PR@2) of 97\%, Mean Reciprocal Rank (MRR) of 0.92 and a Jaccard index of 0.92.

\vspace{-1mm}

\subsubsection{Comparison with Baseline and Ablation Variants}
To assess CausalTrace, we compared it against:
(1) a \textbf{correlation-based root cause ranking}, which identifies candidate causes based on Pearson correlation with the anomalies, and 
(2) a \textbf{CausalTrace variant without KG and ontology grounding}, which omits semantic tool tips, concept filtering, and knowledge-informed edge validation. Results show that the correlation baseline yields high false positives due to spurious associations and fails to capture causal directionality (Jaccard:0.33,MAP@3:44\%,PR@2:51\%,MRR:0.50). The ontology-ablated variant degrades interpretability and occasionally yields invalid causal paths. In contrast, full CausalTrace achieves higher agreement with expert annotations and generates explanations rated as more trustworthy by domain evaluators, with a ROUGE-1 score of 0.91 compared to 0.56 for the variant without ontology (Table \ref{tab:baseline_comparison}).

\begin{table}
\scriptsize
\begin{tblr}{
  rowsep = 1pt, 
  column{even} = {c},
  column{3} = {c},
  column{5} = {c},
  vline{2-6} = {-}{},
  hline{1,6} = {-}{0.08em},
  hline{2} = {1}{},
  hline{2} = {2-6}{0.03em},
}
Method                      & ROUGE-1       & Jaccard       & MAP@3         & PR@2          & MRR           \\
RCA baseline                & –             & 0.33          & 44\%          & 51\%          & 0.5           \\
CausalTrace                 & 0.56          & –             & –             & –             & –             \\
(no KG  ontology)           &               &               &               &               &               \\
\textbf{CausalTrace (full)} & \textbf{0.91} & \textbf{0.92} & \textbf{94\%} & \textbf{97\%} & \textbf{0.92} 
\end{tblr}
\vspace{-2mm}
\caption{Comparison of CausalTrace and baselines}
\label{tab:baseline_comparison}
\vspace{-6mm}
\end{table}

\vspace{-1mm}
\subsubsection{C3AN Principles Evaluation.} We assessed  CausalTrace using the C3AN framework, which defines 14 principles for robust, intelligent, and trustworthy AI systems. A subset of 10 principles was selected based on operational relevance, assessability, and alignment with CausalTrace's capabilities, as in Table \ref{tab:c3an_results}.  To evaluate the system against the C3AN principles, we manually crafted 10 targeted questions per principle using expert guidelines and industry best practices. The ground truth answers were obtained from two sources: (1) domain experts familiar with the manufacturing processes, and (2) official manufacturing manuals and documentation. The ground truth answer was also modeled keeping the C3AN principle in focus\footnote{Questions and results available at: https://shorturl.at/pKWr2}.


To evaluate the system based on the questionnaire, we implemented an \textbf{LLM-as-a-Judge pipeline}. The pipeline consists of state-of-the-art LLMs (\texttt{GPT-4o-mini} and \texttt{LLaMA3-70B-8192}), each provided with three inputs: the question, ground truth, and the CausalTrace-generated answer. Using these inputs as the context, the LLM is prompted to evaluate the generated answer based on the ground truth and provide an integer score from 1 to 5, where 1 signifies complete misalignment with the ground truth and 5 denotes maximum alignment. The LLM is also prompted to explain its score. This enables the model to use more tokens, improving judgment accuracy and reducing bias. The LLM-as-a-Judge framework enabled efficient, low-bias evaluation of system responses. However, since it lacks special domain knowledge, CausalTrace was also assessed by six human evaluators-three experts in the manufacturing sector and three computer scientists. These evaluators scored responses on a scale of 1–5 based on alignment with the ground truth, following the same rubric. (Table \ref{tab:c3an_results}).



The results in Table~\ref{tab:kappa_scores}, indicate that \texttt{GPT-4o-mini} was slightly more conservative in its scoring (\textit{Weighted Cohen’s Kappa} ($\kappa$) mean = 4.19) than \texttt{LLaMA3-70B-8192} (mean = 4.32). In contrast, human evaluators rated more favorably, with an average score of 4.59. The $\kappa$ values further validated the consistency and reliability of these evaluations. Agreement was generally higher within evaluator groups than across them. Specifically, \texttt{GPT-4o-mini} and \texttt{LLaMA3-70B-8192} demonstrated substantial inter-model agreement ($\kappa = 0.56$), whereas their agreement with human evaluators was notably lower ($\kappa = 0.52$ for \texttt{GPT-4o-mini} and $\kappa = 0.58$ for \texttt{LLaMA3-70B-8192}). Human evaluators exhibited moderate internal agreement ($\kappa = 0.58$). These findings suggest that both LLMs and humans tend to align more closely with members of their own group when assessing responses against the ground truth.

\begin{table}[!ht]
\scriptsize
\centering
\begin{tblr}{
  row{1} = {c},
  cell{2}{2} = {c},
  cell{2}{3} = {c},
  cell{2}{4} = {c},
  cell{3}{2} = {c},
  cell{3}{3} = {c},
  cell{3}{4} = {c},
  cell{4}{2} = {c},
  cell{4}{3} = {c},
  cell{4}{4} = {c},
  cell{5}{2} = {c},
  cell{5}{3} = {c},
  cell{5}{4} = {c},
  cell{6}{2} = {c},
  cell{6}{3} = {c},
  cell{6}{4} = {c},
  cell{7}{2} = {c},
  cell{7}{3} = {c},
  cell{7}{4} = {c},
  cell{8}{2} = {c},
  cell{8}{3} = {c},
  cell{8}{4} = {c},
  cell{9}{2} = {c},
  cell{9}{3} = {c},
  cell{9}{4} = {c},
  cell{10}{2} = {c},
  cell{10}{3} = {c},
  cell{10}{4} = {c},
  cell{11}{2} = {c},
  cell{11}{3} = {c},
  cell{11}{4} = {c},
  vline{2-4} = {-}{},
  hline{1-2,12} = {-}{},
  rowsep=1pt, colsep=2pt, 
}
{C3AN Principle, Category and\\~Evaluation Description}                                                                                           & {LLM1 \\Avg.\\Sc.} & {LLM2 \\Avg.\\Sc.} & {Human \\Avg.\\Sc.} \\
{\textbf{Reliability}\textcolor[rgb]{1,0.502,0}{\textbf{[R]:}~}Tested for hallucinations in responses.}                                          & 4                  & 4.5                & 4.2                   \\
{\textbf{Consistency~}\textcolor[rgb]{1,0.502,0}{\textbf{[R]}:~}Assessed stability using paraphrased \\ queries and comparing output similarity.} & 3.5                & 4.5                & 4.4                 \\
{\textbf{Abstraction}\textcolor{darkgreen}{\textbf{[I]}:~}Evaluated how low-level data are mapped \\to high-level concepts in knowledge graph and ontology}                     & 4                  & 3.3                & 4.6                 \\
{\textbf{Causality}\textcolor{darkgreen}{\textbf{[I]}:~}Tested via causal discovery, total\\effects, root cause analysis, and counterfactuals.}                                   & 4.9                & 4.4                & 4.7                 \\
{\textbf{Reasoning}\textcolor{darkgreen}{\textbf{[I]}:~}  Measured the system's ability to\\logically infer conclusions.}                                                           & 3.8                & 4.5                & 4.6                 \\
{\textbf{Planning}\textcolor{darkgreen}{\textbf{[I]}:~}  Evaluated multi-agent workflows from\\data input to final output.}                                                         & 4                  & 5                  & 4.6                 \\
{\textbf{Grounding}\textcolor{blue}{\textbf{[T]:}~}Checked if answers used context\\from knowledge graph and ontology.}                           & 4.4                  & 4.6                  & 4.6                   \\
{\textbf{Attribution}\textcolor{blue}{\textbf{[T]}:~}Verified traceability of decisions\\using agent logs and contextual sources.}                 & 4.5                & 4.5                & 4.2                   \\
{\textbf{Interpretability}\textcolor{blue}{\textbf{[T]}:}~Users could inspect model\\settings, selected features, and causal graphs.}             & 4                  & 3.3                & 4.5                 \\
{\textbf{Explainability\textcolor{blue}{[T]:~}}Judged clarity of explanations\\for and causal outputs.}                        & 5                  & 5                  & 5                   
\end{tblr}
\vspace{-2mm}
\caption{Average performance on selected C3AN principles, evaluated by two LLMs (\texttt{GPT-4o-mini} and \texttt{LLaMA3-70B-8192}) and six human evaluators. Categories: \textcolor{orange}{R}= Robustness, \textcolor{darkgreen}{I}= Intelligence, \textcolor{blue}{T}= Trustworthiness.}
\label{tab:c3an_results}
\vspace{-4mm}
\end{table}

\vspace{-3mm}
\begin{table}[!ht]
\scriptsize
\begin{tabular}{>{\hspace{2pt}}l@{\hskip 8pt}c@{\hskip 4pt}c@{\hskip 4pt}c@{\hskip 4pt}c@{\hskip 4pt}c@{\hskip 4pt}c@{\hskip 4pt}c@{\hskip 4pt}c}
\toprule
 & LLM1 & LLM2 & E1 & E2 & E3 & E4 & E5 & E6 \\
\midrule
LLM1 & \cellcolor{green!50}\makebox[3em]{1} 
     &  &  &  &  &  &  &  \\
     
LLM2 & \cellcolor{green!40}\makebox[3em]{0.61} 
     & \cellcolor{green!50}\makebox[3em]{1} 
     &  &  &  &  &  &  \\
     
E1   & \cellcolor{green!70}\makebox[3em]{0.68} 
     & \cellcolor{green!40}\makebox[3em]{0.68} 
     & \cellcolor{green!50}\makebox[3em]{1} 
     &  &  &  &  &  \\
     
E2   & \cellcolor{yellow!50}\makebox[3em]{0.48} 
     & \cellcolor{green!40}\makebox[3em]{0.56} 
     & \cellcolor{green!40}\makebox[3em]{0.56} 
     & \cellcolor{green!50}\makebox[3em]{1} 
     &  &  &  &  \\
     
E3   & \cellcolor{green!40}\makebox[3em]{0.50} 
     & \cellcolor{yellow!50}\makebox[3em]{0.45} 
     & \cellcolor{green!70}\makebox[3em]{0.61} 
     & \cellcolor{green!40}\makebox[3em]{0.53} 
     & \cellcolor{green!50}\makebox[3em]{1} 
     &  &  &  \\
     
E4   & \cellcolor{green!70}\makebox[3em]{0.61} 
     & \cellcolor{green!40}\makebox[3em]{0.59} 
     & \cellcolor{green!70}\makebox[3em]{0.64} 
     & \cellcolor{green!70}\makebox[3em]{0.86} 
     & \cellcolor{green!40}\makebox[3em]{0.58} 
     & \cellcolor{green!50}\makebox[3em]{1} 
     &  &  \\
     
E5   & \cellcolor{red!30}\makebox[3em]{0.33} 
     & \cellcolor{green!40}\makebox[3em]{0.66} 
     & \cellcolor{green!40}\makebox[3em]{0.52} 
     & \cellcolor{green!40}\makebox[3em]{0.57} 
     & \cellcolor{yellow!50}\makebox[3em]{0.54} 
     & \cellcolor{green!40}\makebox[3em]{0.52} 
     & \cellcolor{green!50}\makebox[3em]{1} 
     &  \\
     
E6   & \cellcolor{green!70}\makebox[3em]{0.52}
     & \cellcolor{yellow!50}\makebox[3em]{0.46}
     & \cellcolor{green!40}\makebox[3em]{0.51}
     & \cellcolor{green!40}\makebox[3em]{0.57}
     & \cellcolor{green!70}\makebox[3em]{0.86}
     & \cellcolor{yellow!50}\makebox[3em]{0.49}
     & \cellcolor{yellow!50}\makebox[3em]{0.37}
     & \cellcolor{green!50}\makebox[3em]{1} \\
\bottomrule
\end{tabular}
\vspace{-2mm}
\caption{Inter-annotator agreements ($\kappa$) (range: –1 to 1). Green: perfect agreement, yellow: moderate, red: low. LLM1: \texttt{GPT-4o-mini}, LLM2: \texttt{LLaMA3-70B-8192}, E1–E6: Human evaluators.}
\label{tab:kappa_scores}
\vspace{-8mm}
\end{table}

\section{Pathway To Deployment}
\begin{figure}[!ht]
\vspace{-2mm}
\includegraphics[width=0.999\linewidth]{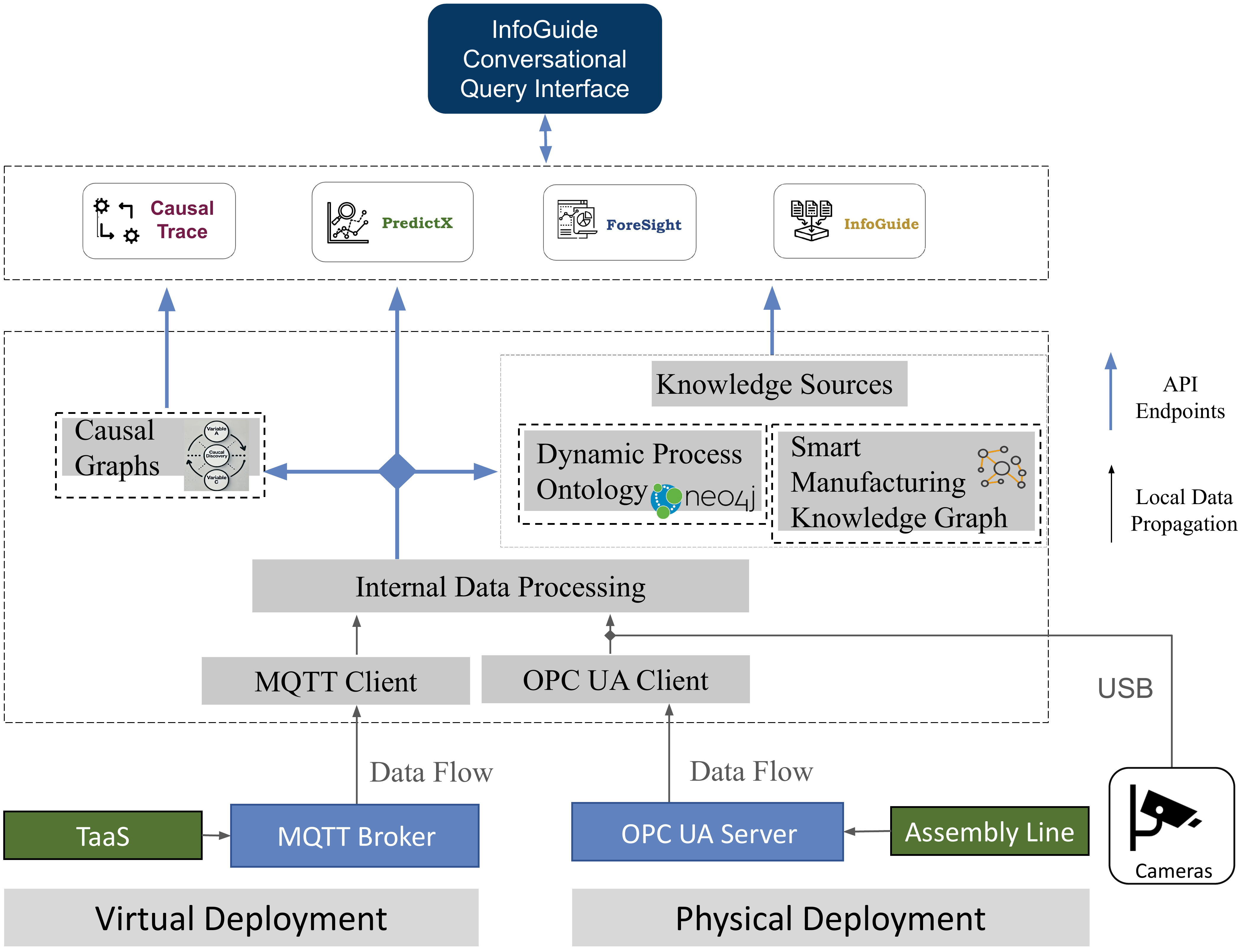}
\vspace{-6mm}
\caption{End-to-end deployment pipeline of SmartPilot with virtual and real-world integration}
  \label{fig:DepArch}
  \vspace{-7mm}
\end{figure}

\vspace{-1mm}
CausalTrace is integrated as a core agent in SmartPilot, which is currently under deployment in the FF Lab. The full deployment pipeline proceeds through two primary phases: virtual and real-world deployment (Figure~\ref{fig:DepArch}). 
In the initial phase, SmartPilot is virtually deployed using the Testbed as a Service (TaaS) framework \cite{mccormick2025testbed}, which emulates the robotic assembly line operations.
TaaS functions as a data replay mechanism, publishing data from a pre-existing dataset one timestamp at a time, mimicking the behavior of live data streams. In this setting, SmartPilot subscribes to specific MQTT (Message Queuing Telemetry Transport) topics defined by TaaS and processes the incoming data streams as if operating in a live production environment. This virtual prototyping setup offers a controlled and repeatable environment for evaluating system behavior under a range of operational conditions.

Following successful virtual validation, SmartPilot progresses to real-world deployment. The process begins with configuring the OPC UA Server at the assembly line level.  Subsequently, analog sensor data is streamed via the OPC UA Server, while image data is acquired directly from cameras positioned in the lab. This phase encompasses internal data processing tasks, including structuring raw inputs and developing a robust data ingestion pipeline. Next, the knowledge graph, dynamic process ontology, and previously generated in-memory causal graphs are instantiated via Neo4j, RDF, and HTTP SSE, respectively, and integrated with the data infrastructure. Finally, leveraging the ingested timeseries signals, visual data, and semantic insights from knowledge sources, SmartPilot’s four specialized agents operate in real time to perform anomaly detection, production forecasting, question answering, and causal analysis. The current deployment architecture is optimized for minimal latency in data acquisition and processing, as demonstrated in \cite{elkalach2025real}. However, its centralized nature presents scalability limitations as system complexity increases. To address this, future iterations of the system will transition to a decentralized processing model. Furthermore, the direct connection to cameras imposes spatial constraints, as the system responsible for image ingestion must remain within the physical range of the camera cabling.

In the second iteration of the deployment, the OPC UA Server and direct camera connections are to be replaced by a public MQTT broker, enabling SmartPilot to be hosted in a publicly accessible environment. This revised architecture introduces an additional layer beneath the data processing stage: a data publishing module that streams data from the local servers and cameras to the MQTT broker. SmartPilot is also publicly hosted\footnote{\url{https://aiisc.ai/smartpilot/}} to enable community access, leveraging both in-house and community-driven abstractions for interfacing with its back-end services. This setup enables global real-time access, improving scalability and accessibility, but it introduces latency, which we plan to address with edge computing and asynchronous communication.

\vspace{-2mm}
\section{Conclusion and Further Work}
\vspace{-1mm}

We introduced CausalTrace, a neurosymbolic causal analysis agent for intelligent manufacturing. Integrated with knowledge sources, it supports interpretable and trustworthy reasoning over complex industrial data. Evaluated via the C3AN framework, CausalTrace demonstrates strong alignment with robustness, intelligence, and trustworthiness. Its phased deployment from controlled virtual prototyping to ongoing live integration shows practical viability and scalability potential. Future work will extend capabilities such as intervention planning, continual causal graph learning, and safety-aware, instruction-following features. These enhancements aim to further optimize performance and adaptability in diverse operational environments.

\section{Acknowledgment}
This work was supported in part by NSF grant \#2119654, “RII Track 2 FEC: Enabling Factory to Factory (F2F) Networking for Future Manufacturing”.

\bibliography{aaai2026}
\clearpage
\onecolumn

\section{Appendix}

\section{Appendix A: CausalTrace Competency Questions}
\label{sec:appA}

This section presents a structured set of competency questions designed to evaluate the capabilities of the CausalTrace across key reasoning dimensions. These questions are grounded in the example causal graph \( G = \{A \rightarrow B,\, B \rightarrow C,\, D \rightarrow B\} \), which is assumed to be generated by the CausalTrace and span four core categories: (1) Causal Graph Structure, (2) Causal Reasoning, (3) Root Cause Analysis, and (4) Causal Discovery. The goal is to assess the agent’s ability to interpret, reason over, and explain causal relationships in response to natural language queries.

\subsubsection{Questions about Causal Graph Structure}
\begin{enumerate}
    \item Does A cause B?
    \item Does B cause A?
    \item Does A cause C?
    \item Is there a causal relation between A and B (or between B and A)?
    \item Is there a causal relation between A and D?
    \item What variables have a direct causal effect on B (i.e., causal parents of B)?
    \item Is there a causal relation between A and Z?
\end{enumerate}

\subsubsection{Questions about Causal Reasoning}
\begin{enumerate}
    \item What is the strength of the causal relation between A and B?
    \item Which variable has the strongest causal effect on B?
    \item If the value of A were set to \( x \), what would be the effect on B?
\end{enumerate}

\subsubsection{Questions about Root Cause Analysis}
\begin{enumerate}
    \item What is the most likely root cause of the anomalous value of variable B?
    \item What are the three most likely root causes of the anomalous value of variable B, ranked in descending order?
    \item Is A a likely root cause of the anomalous value of variable B?
    \item Which is more likely to be the root cause: A or D?
    \item Why is D not considered a likely root cause of the anomalous value of variable B?
\end{enumerate}

\subsubsection{Questions about Causal Discovery}
\begin{enumerate}
    \item What algorithm was used to learn the causal graph?
    \item What is the stability score of the edge \( A \rightarrow B \)?
    \item Which edges in the graph are considered the least reliable?
    \item How many bootstrap iterations were used during causal discovery?
\end{enumerate}

\end{document}